\documentclass[a4paper]{article}

\usepackage{fullpage} 
\usepackage{parskip} 
\usepackage{tikz} 
\usepackage{amsmath}
\usepackage{hyperref}
\usepackage{subfig}

\usepackage{graphicx}             
\usepackage{moreverb}             

\usepackage{graphicx}
\usepackage{amsmath,amssymb} 
\usepackage{color}
\usepackage{blindtext}
\usepackage{tabto, hyperref}
\usepackage{subfig, array}
\usepackage{blindtext, amsmath}
\usepackage{graphicx, tabto, hyperref}
\usepackage{subfig}
\usepackage{dblfloatfix}
\usepackage{graphics}
\usepackage{fontenc}
\usepackage{txfonts}
\usepackage{pifont}
\usepackage{color}
\usepackage{hyperref}
\usepackage{booktabs}
\usepackage{multirow}
\usepackage{float}
\usepackage{siunitx, array}
\usepackage{authblk}
\usepackage[linesnumbered,ruled]{algorithm2e}
\newcolumntype{P}[1]{>{\centering\arraybackslash}p{#1}}

\title{CoCoNet: A Collaborative Convolutional Network}
\author[1,2]{Tapabrata Chakraborti}
\author[1]{Brendan McCane}
\author[1]{Steven Mills}
\author[2]{Umapada Pal}

\affil[1]{Dept. of Computer Science, University of Otago, NZ}
\affil[2]{CVPR Unit, Indian Statistical Institute, India}

\begin{document}

\maketitle

\begin{abstract}

We present an end-to-end deep network for fine-grained visual categorization called Collaborative Convolutional Network (CoCoNet). The network uses a collaborative layer after the convolutional layers to represent an image as an optimal weighted collaboration of features learned from training samples as a whole rather than one at a time. This gives CoCoNet more power to encode the fine-grained nature of the data with limited samples. We perform a detailed study of the performance with 1-stage and 2-stage transfer learning. The ablation study shows that the proposed method outperforms its constituent parts consistently. CoCoNet also outperforms few state-of-the-art competing methods. Experiments have been performed on the fine-grained bird species classification problem as a representative example, but the method may be applied to other similar tasks. We also introduce a new public dataset for fine-grained species recognition, that of Indian endemic birds and have reported initial results on it.

\end{abstract}

\section{Introduction}




Deep convolutional networks have proven to be effective in classifying base image categories with sufficient generalization when trained with a large dataset. However, many real life applications of significance [1] may be characterized by fine-grained classes and limited availability of data, like endangered species recognition or analysis of biomedical images of a rare pathology. In such specialized problems, it is challenging to effectively train deep networks that are, by nature, data hungry. A case in point is that of fine-grained endangered species recognition [2], where besides scarcity of training data, there are further bottlenecks like subtle inter-class object differences compared to significant randomized background variation both between and within classes. This is illustrated in Fig. 1. Added to these, is the presence of the ``long tail" problem [3], that is, significant imbalance in samples per class (the frequency distribution of samples per class has long tail).

Transfer learning is a popular approach to train on small fine-grained image datasets with limited samples [4]. The ConvNet architecture is trained first on a large benchmark image dataset (e.g. ImageNet) for the task of base object recognition. The network is then fine-tuned on the smaller target dataset for fine-grained recognition. Since the target dataset is small, there is an increased chance of overtraining. On the other hand, if the dataset has fine-grained objects with varying backgrounds, this can cause difficulty in training convergence. This makes the optimal training of the dataset challenging [3]. In case of small datasets with imbalanced classes, the problem is compounded by the probability of training bias in favour of larger classes. A few specialized deep learning methods have been proposed in recent times to cater to these issues, like low-shot/zero-shot learning [5] for small datasets and multi-staged transfer learning [4] for fine-grained classes. In spite of these advances, deep learning of small fine-grained datasets remains one of the open popular challenges of machine vision [6][7].

Recently, Chakraborti \emph{et al}. [8] demonstrated that collaborative filters can effectively represent and classify small fine-grained datasets. Collaborative filters have been the method of choice for recommender systems [19] and have recently also been used in vision systems like face recognition [9]. Cai \emph{et al}. [10] have recently shown that some modern versions of Collaborative Representation Classifiers (CRC) give better performance with CNN learned features from a pre-trained ConvNet compared to softmax based classification layer [10]. 

The intuition why collaborative filtering works well to represent fine-grained data is as follows. Collaborative representation classifiers (CRC) [9] represent the test image as an optimal weighted average of all training images across all classes. The predicted label is the class having least residual. This inter-class collaboration for optimal feature representation is distinct from the traditional purely discriminative approach. Thus the collaborative scheme not only takes advantages of differences between object categories but also exploits the similarities of fine-grained data. Other advantages are that CRC is analytic since it has a closed form solution and is time efficient since it does not need iterative or heuristic optimization. It is also a general feature representation-classification scheme and thus most popular feature descriptors and ensembles thereof are compatible with it. The present work advances the state-of-the-art by encoding the collaborative loss function into a deep CNN model. The contributions of this paper are two-fold.


\textbf{1. Collaborative ConvNet (CoCoNet)}: The proposed method fine-tunes a pre-trained deep network through a collaborative representation classifier in an end-to-end fashion. This establishes a protocol for multi-stage transfer learning of fine-grained data with limited samples. We test it for fine-grained bird species recognition as an example, but the same may be applied to other similar tasks.
    
    
\textbf{2. Indian Birds dataset}: We introduce IndBirds, a new fine-grained image benchmark of Indian endemic birds. It currently has 800 images of 8 classes (100 images per class). All experiments have been repeated on the new dataset and results are presented.

\section{Collaborative Convolutional Network}

We first present a brief description of the original collaborative representation classifier (CRC) [9] and then the proposed Collaborative ConvNet (CoCoNet).

\subsection{Collaborative Representation Classifier (CRC)}

Consider a training dataset with images in the feature space as $X=[X_1,\dots,X_c]\in \varmathbb{R}^{d \times n}$ where $n$ is the total number of samples over $c$ classes and $d$ is the feature dimension per sample. Thus $X_i \in \varmathbb{R}^{d \times n_i}$ is the feature space representation of class $i$ with $n_i$ samples such that $\sum_{i=1}^{c} n_i = n$.

The CRC model reconstructs a test image in the feature space $\vec{y} \in \varmathbb{R}^d$ as an optimal collaboration of all training samples, while at the same time limiting the size of the reconstruction parameters, using the \emph{l}2 regularisation term $\lambda$.

The CRC cost function is given by: 
\begin{equation}
J(\alpha,\lambda)=\|\vec{y}-X\alpha\|_2^2+\lambda\|\alpha\|_2^2
\end{equation}                  
where $\hat{\alpha}=[\hat{\alpha}_1,\dots,\hat{\alpha}_c]\in\varmathbb{R}^N \mid \hat{\alpha}_i\in\varmathbb{R}^{n_i}$ is the reconstruction matrix corresponding to class $i$.

\noindent A least-squares derivation yields the optimal solution for $\alpha$ as:
\begin{equation}
\hat{\alpha}=(X^TX+ \lambda I)^{-1}X^T\vec{y}
\end{equation}     

\noindent The representation residual of class $i$ for test sample $\vec{y}$ can be calculated as: 
\begin{equation}
r_i(\vec{y})=\frac{\|\vec{y}-X_i\hat{\alpha}_i\|_2^2}{\|\hat{\alpha}_i\|_2^2} \ \forall i \in {1,\dots,c}
\end{equation}

\noindent The final class of test sample $\vec{y}$ is thus given by
\begin{equation}
C(\vec{y})=\text{arg}\,\min\limits_i\, r_i(\vec{y})
\end{equation}  

\noindent The optimal value of $\lambda$ is determined using gradient descent.

\subsection{Collaborative ConvNet (CoCoNet)}

CoCoNet gives a collaborative cost which is back propagated through an end-to-end model. The training set is divided into two sections $p1$ and $p2$, having $m$ and $n$ images respectively randomly selected with equal representation across classes. 

Let $x$ be the $d\times1$ feature vector of one image in $p1$, such that the feature matrix for $p1$ is $X$ of dimension $d \times m$. Let $y$ be the $d\times1$ feature vector of one image in $p2$, such that the feature matrix for $p2$ is $Y$ of dimension $d \times n$. 

\begin{figure*}[]
\centering

\subfloat{\includegraphics[width=5in,height=2in]{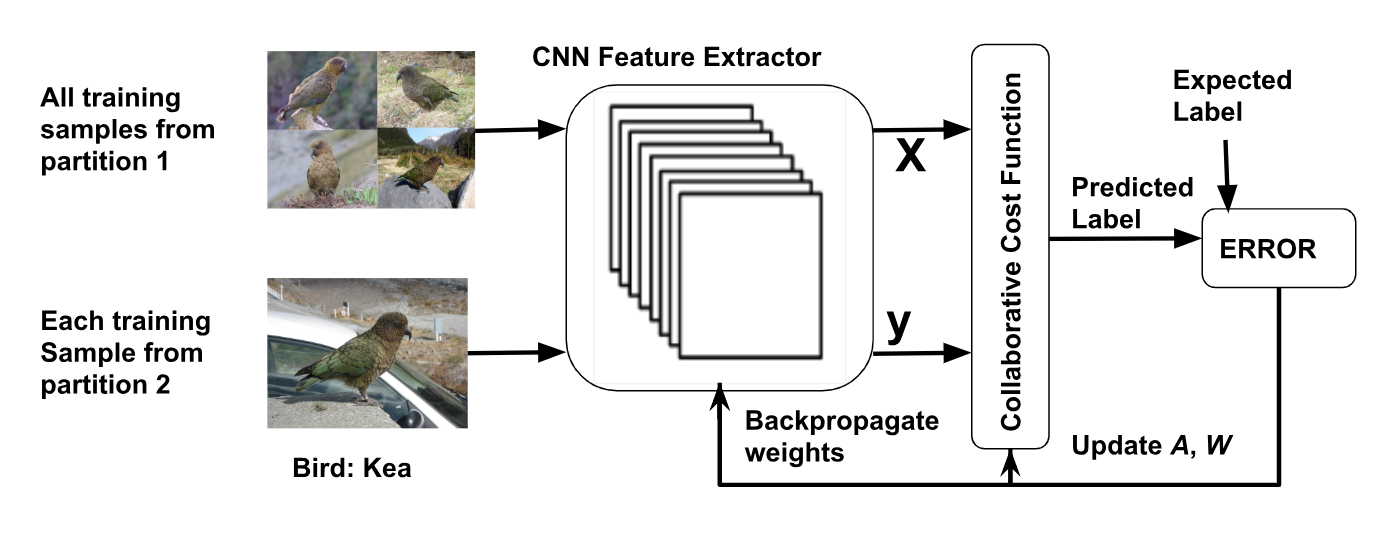}}

\caption{Architecture of the proposed Collaborative ConvNet (CoCoNet). Training samples are divided into two partitions which collaborate to represent fine-grained patterns. The collaborative cost function generates the error that optimizes its own parameters as well as fine-tune the network weights in an end-to-end manner.} 
\end{figure*}

The collaborative cost function is given by:

\begin{equation}
P \ (A,W,X) =\|(Y-XA)W\|_2^2+\lambda\|A\|_2^2+\gamma\|W\|_2^2
\end{equation}

The collaborative reconstruction matrix $A$ is thus of dimension $m \times n$. The goal is to find an optimal feature representation of each sample in $p2$ with respect to the ``training" images in $p1$ via a representation vector $\vec{a}_i \in A$.

The weight matrix $W$ is used to compensate for imbalance of classes and each of its elements is initialised with a weight proportional to the size of the class to which the corresponding feature vector in $Y$ belongs. $W$ counteracts the imbalance in classes as a penalty term for larger classes by increasing the cost. $W$ is of dimension $n \times 1$.

After finding the initial optimal $A$ through least squares, it is then updated along with the weight matrix $W$ through partial derivatives. The gradient of the feature matrix $X$ is then used to update the CNN weights through back-propagation as presented in Algorithm 1. \\

Least squares minimization gives the initial optimal value of $A$ as:

\begin{equation}
\hat{A}=\Big[X^TXW^TW+\lambda I\Big]^{-1}X^TYWW^T
\end{equation}

Fix $\big\{A,X\big\}$, update $W$:
\begin{equation}
\frac{\partial P}{\partial W}=-(Y-XA)^T(Y-XA)W+\gamma W
\end{equation}

Fix $\big\{W,X\big\}$, update $A$:
\begin{equation}
\frac{\partial P}{\partial A}=-X^T(Y-XA)WW^T+\lambda A
\end{equation}

Fix $\big\{W,A\big\}$, update $X$:
\begin{equation}
\frac{\partial P}{\partial X}=-(Y-XA)WW^TA^T
\end{equation}

Once all the partial derivatives are obtained, CNN weights are updated by standard back-propagation of gradients for each batch in $P1$ and $P2$. The training iterations are continued until the error stabilizes. A schematic of the CoCoNet architecture is presented in Fig. 1. \\

\begin{algorithm}[]
\SetAlgoLined
 \textbf{Initiate} weight matrix $W$ proportional to class size \;
 \textbf{Split} the training set into two parts $P1$ and $P2$ \;
 \textbf{Extract} feature matrix $X$ of $P1$ through CNN section. \;
 \textbf{Find} initial optimal reconstruction matrix $A$ by eqn. 6. \;
 \For{each sample in $P2$}{
 Fix $\big\{A,X\big\}$, update $W$ by eqn. 7 \;
 Fix $\big\{W,X\big\}$, update $A$ by eqn. 8 \;
 \For{each sample in $P1$}
 {Back-propagate to update weights using eqn. 9 \;}
  }

 \caption{Training with CoCoNet}
\end{algorithm}


\subsection{Reducing computation cost through SVD.} The optimal representation weight matrix $\hat{A}$ from eqn 6. has the term $(X^TXW^TW + \lambda I)^{-1}$, where $X$ is of dimension $d \times m$. Here $d$ is the dimension of the descriptor and $m$ is the total number of data points in the partition $P1$ of training data. This poses the problem of high computation cost for large datasets ($m$ is large). So we use singular value decomposition (SVD) to reduce the matrix inverse computation to dimension $d \times d$, so as to make it independent of dataset size. This is a crucial modification needed for applications like image retrieval from large unlabeled or weakly labeled image repositories. \\

If we take the singular value decomposition (SVD) of $X^T$, we can factor $X^TX$ as:

\begin{equation}
X^TX = (USV^T)^T USV^T
= V S^T U^T U S V^T
= V (S^2) V^T
\end{equation}

Since $S$ only has $d$ non-zero singular values, we can truncate $S^TS$ and $V$ to be smaller matrices. So $V$ is $N \times d$, $S$ is $d \times d$ and $V^T$ is $d \times N$. Also note that since $W$ is of dimension $n \times 1$, $W^TW$ comes out as a scalar value in eqn. 6, which is absorbed in $S$ to have $\hat{S}$.

Using the Woodbury matrix inverse identity [18], the inverse term becomes $(V\hat{S}^2V^T + \lambda I)^{-1}$ which can be represented as:
\begin{equation}
 \frac{1}{\lambda} + \frac{1}{\lambda^2}V (\hat{S}^{-1} + \frac{1}{\lambda}V^TV)^{-1} V^T
= \frac{1}{\lambda} + \frac{1}{\lambda^2} V(\hat{S}^{-1} + \frac{1}{\lambda} I)^{-1} V^T
\end{equation}

Note that the inverse term $(\hat{S}^{-1} + \frac{1}{\lambda} I)^{-1}$ is only $d \times d$, so it scales to many data points.

\subsection{Enhanced Learning by CoCoNet}

CoCoNet uses the collaborative cost function in an end-to-end manner. So we do not have the fully connected, energy loss function and softmax layers. The CNN extracts features and feeds it to the collaborative layer. The collaborative cost function estimates error, updates its own weights as well as feeds the error back to the CNN. The error and gradients are then back propagated through the CNN to update the weights. So CoCoNet is different from just cascading a CNN based feature learner with a collaborative filter, because the weights are not updated in latter in an end-to-end fashion. For the same dataset and same number of given samples, the collaborative filter represents all samples together as an augmented feature vector. Thus after the error is calculated, the error gradients are then back propagated. This collaborative representation is not just the augmented feature matrix with all samples, it is also optimised by the collaborative filter. This adds an additional level of optimisation besides the CNN learned features, weights and tuned parameters.

\section{Experiments and Results}

\subsection{Datasets} 

Five benchmark image datasets are used in this work for pre-training and fine-tuning in total.

\textbf{ImageNet} [11] has about 1.4 million image categories and about 22k indexed sysnet as of 2017. It has been used for pre-training the networks as base category classifiers. Then for transfer learning, the following bird species recognition datasets have been used.

\textbf{NABirds} [12] is a fine-grained North American bird species recognition dataset developed by Cornell-UCSD-CalTech collaboration and maintained by the Cornell Lab of Ornithology. It is continually updated and at the time of use for this work had 555 classes and 48562 images. Due to the large number of images present in this dataset, it may be used for training a deep network from scratch as well as for transfer learning.

\textbf{CUB 200-2011} [13] dataset contains 11,788 images of 200 bird species. The main challenge of this dataset is considerable variation and confounding features in background information compared to subtle inter-class differences in birds.

\textbf{IndBirds} is a new bird species recognition benchmark compiled as part of this work by the Indian Statistical Institute and University of xxxx, NZ. The dataset contains images of 8 species of Indian endemic birds with around 100 images per class, collected from web repositories of birders and citizen scientists. The dataset is available for academic use from the lead author. Sample images of each class are presented in Fig. 2.

\textbf{NZBirds} [14] is a small benchmark dataset of fine-grained images of NZ endemic birds, many of which are endangered. Currently it contains 600 images of 30 NZ birds and has been compiled by University of Otago in collaboration with The National Museum of NZ (Te Papa), the Department of Conservation (DOC) and the Ornithological Society of NZ (Birds NZ).

\begin{figure}[h!]
\centering

\subfloat{\includegraphics[width=0.66in,height=0.55in]{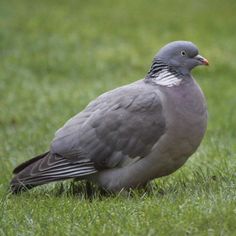}}
\hfil
\subfloat{\includegraphics[width=0.66in,height=0.55in]{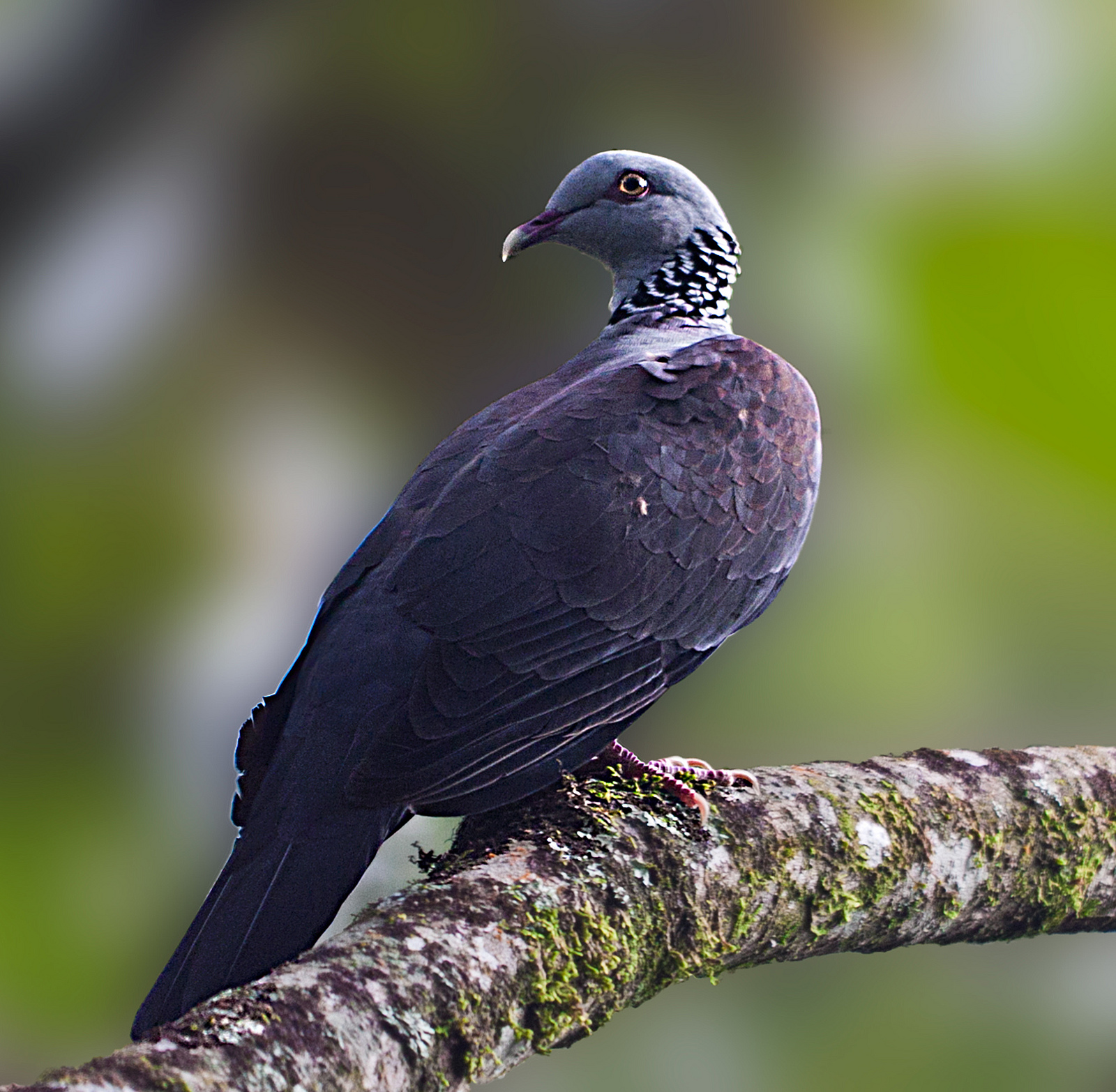}}
\hfil
\subfloat{\includegraphics[width=0.66in,height=0.55in]{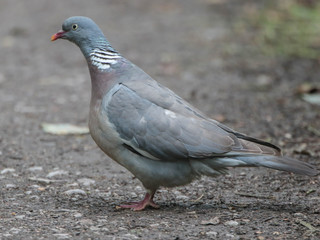}}
\hfil
\subfloat{\includegraphics[width=0.66in,height=0.55in]{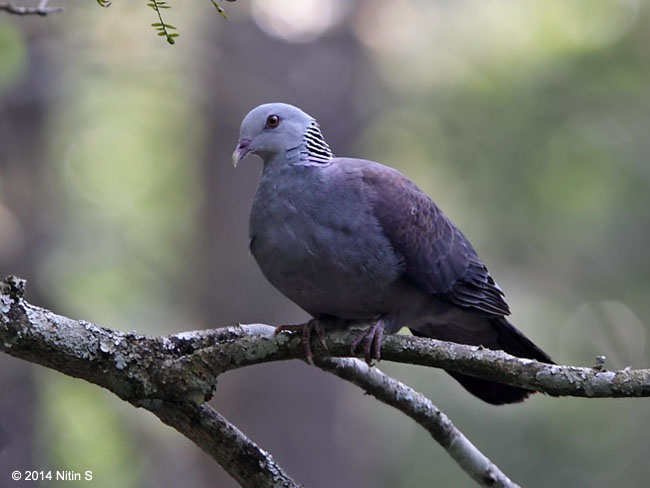}}
\hfil
\subfloat{\includegraphics[width=0.66in,height=0.55in]{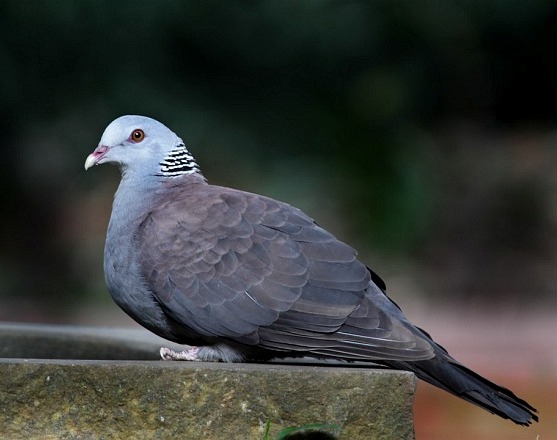}}
\\
Nilgiri Wood Pigeon
\\
\subfloat{\includegraphics[width=0.66in,height=0.55in]{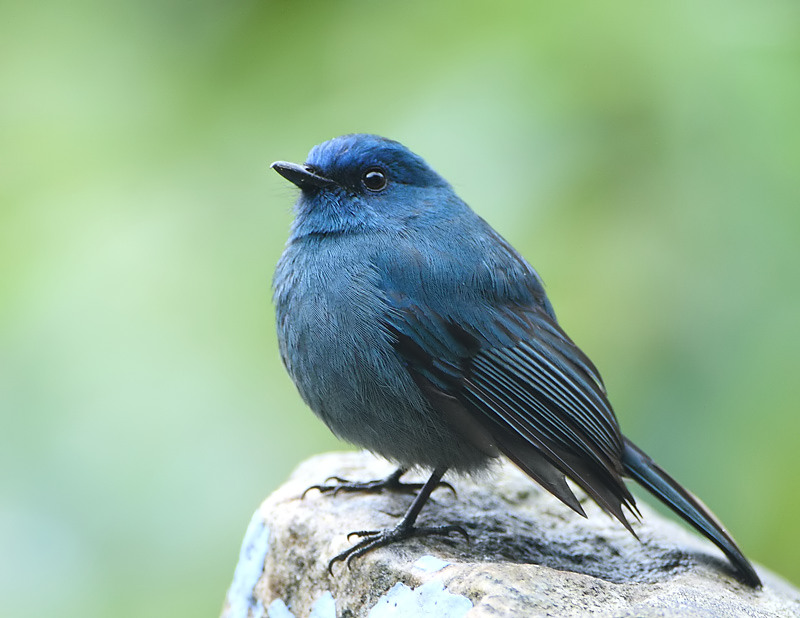}}
\hfil
\subfloat{\includegraphics[width=0.66in,height=0.55in]{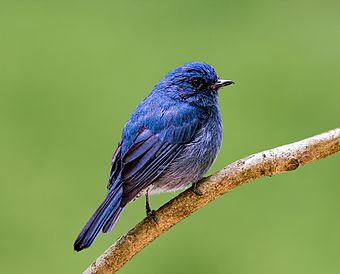}}
\hfil
\subfloat{\includegraphics[width=0.66in,height=0.55in]{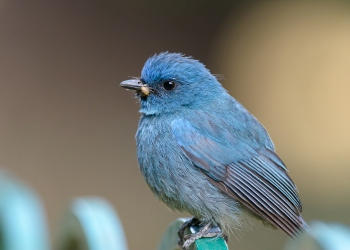}}
\hfil
\subfloat{\includegraphics[width=0.66in,height=0.55in]{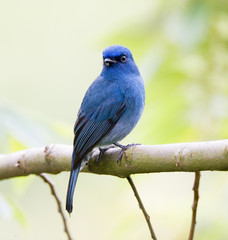}}
\hfil
\subfloat{\includegraphics[width=0.66in,height=0.55in]{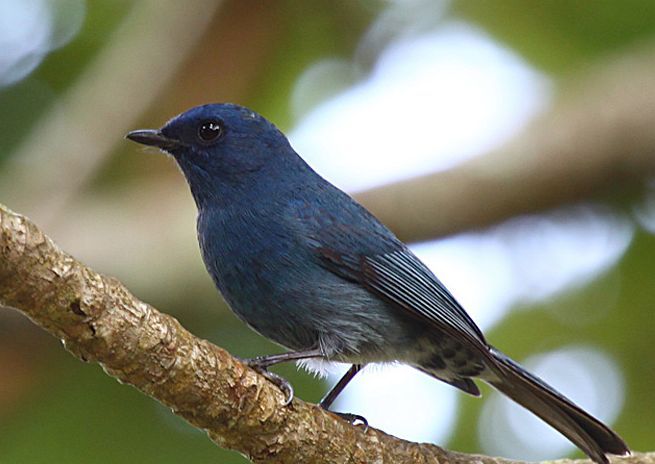}}
\\
Nigiri Fly Catcher
\\
\subfloat{\includegraphics[width=0.66in,height=0.55in]{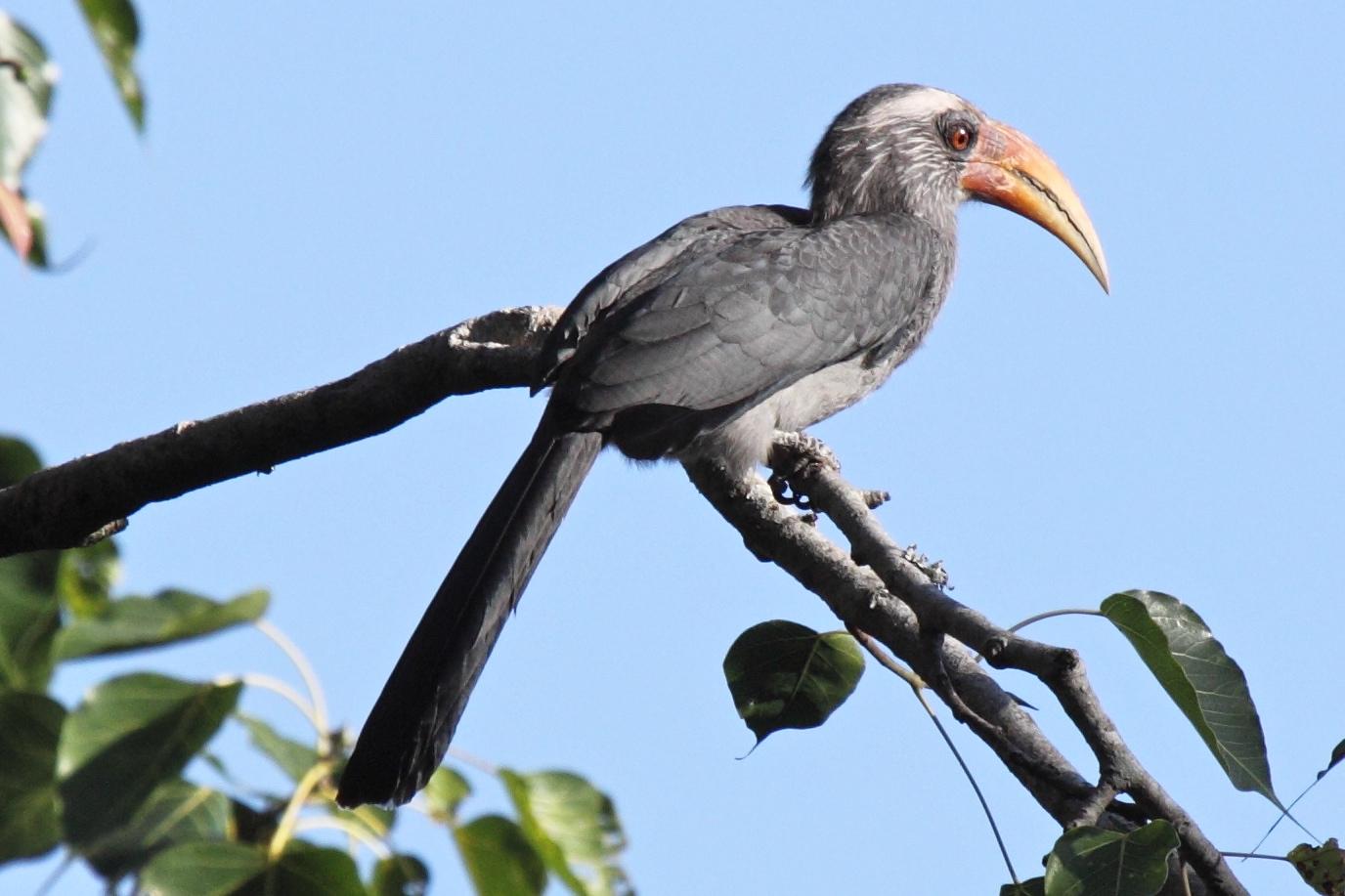}}
\hfil
\subfloat{\includegraphics[width=0.66in,height=0.55in]{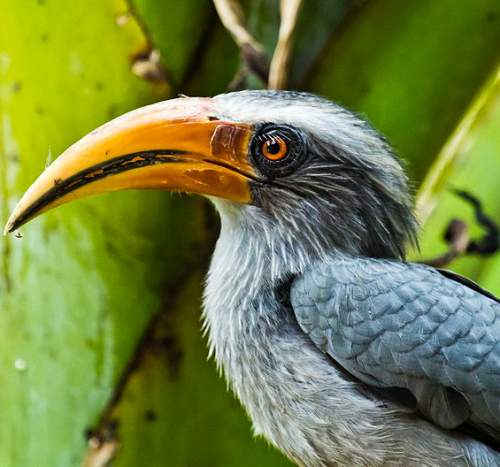}}
\hfil
\subfloat{\includegraphics[width=0.66in,height=0.55in]{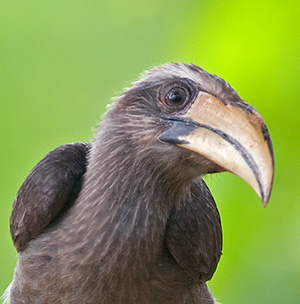}}
\hfil
\subfloat{\includegraphics[width=0.66in,height=0.55in]{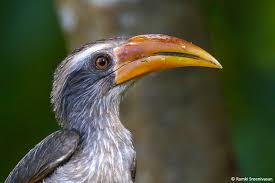}}
\hfil
\subfloat{\includegraphics[width=0.66in,height=0.55in]{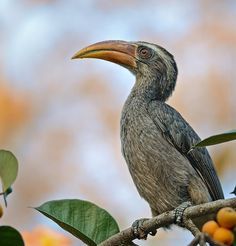}}
\\
Malabar Grey Hornbill
\\
\subfloat{\includegraphics[width=0.66in,height=0.55in]{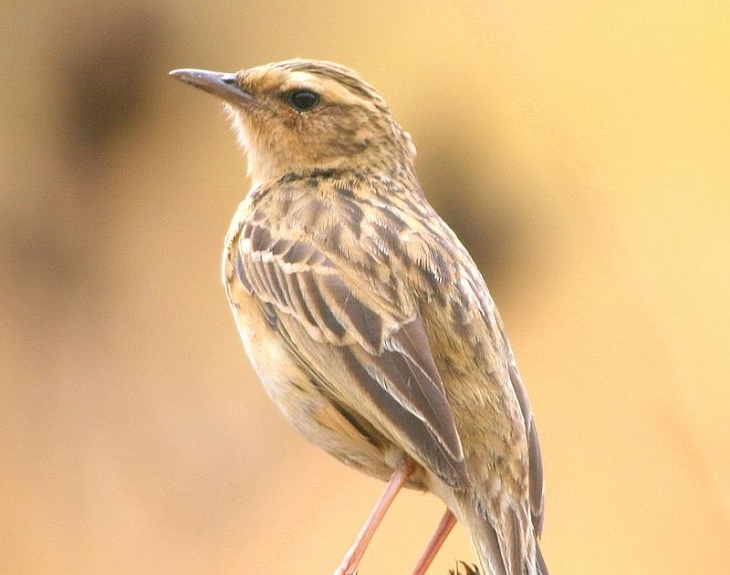}}
\hfil
\subfloat{\includegraphics[width=0.66in,height=0.55in]{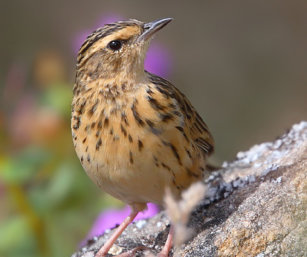}}
\hfil
\subfloat{\includegraphics[width=0.66in,height=0.55in]{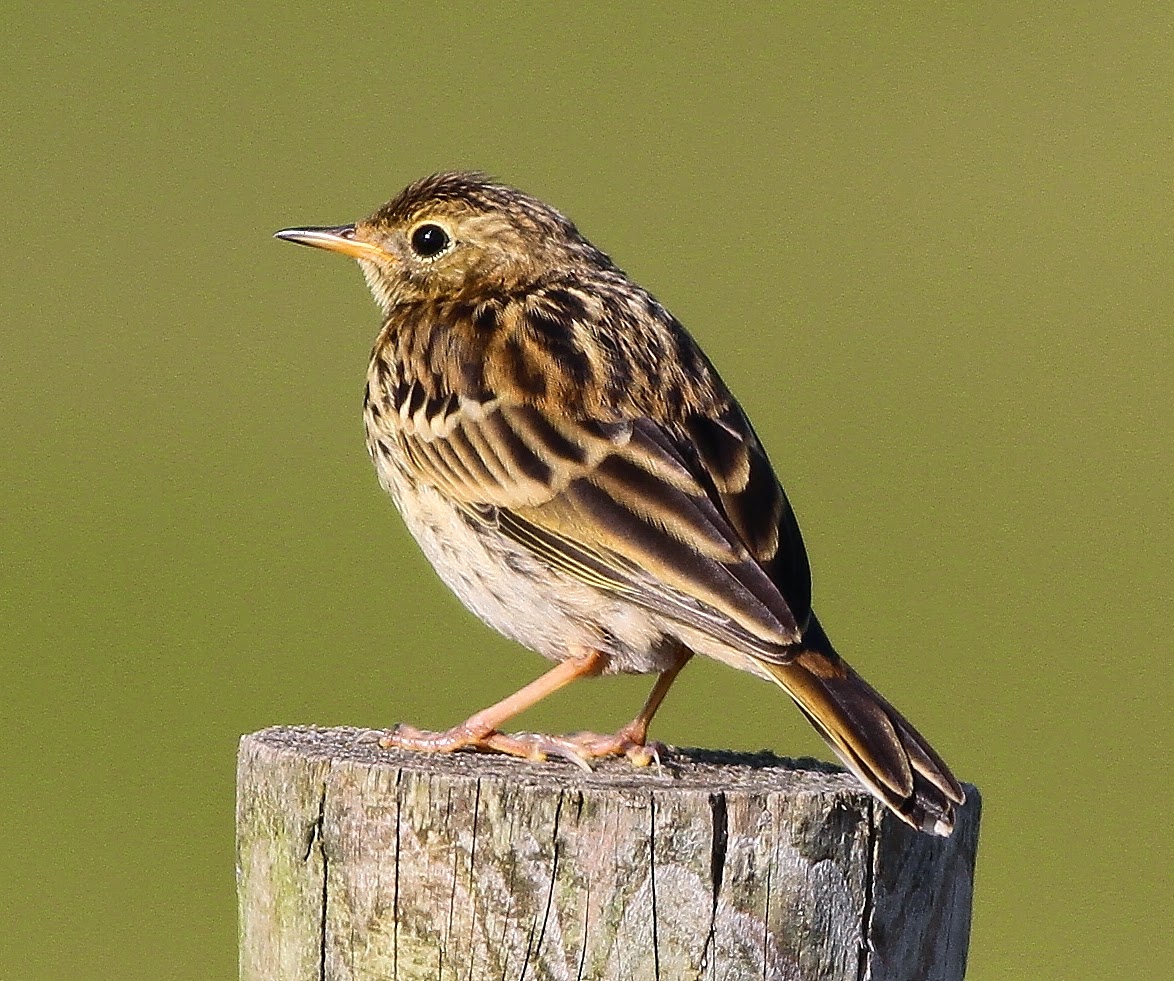}}
\hfil
\subfloat{\includegraphics[width=0.66in,height=0.55in]{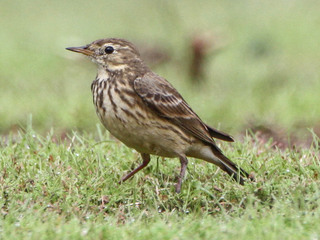}}
\hfil
\subfloat{\includegraphics[width=0.66in,height=0.55in]{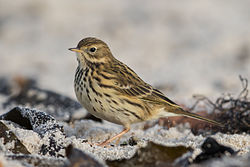}}
\\
Nilgiri Pipit
\\
\subfloat{\includegraphics[width=0.66in,height=0.55in]{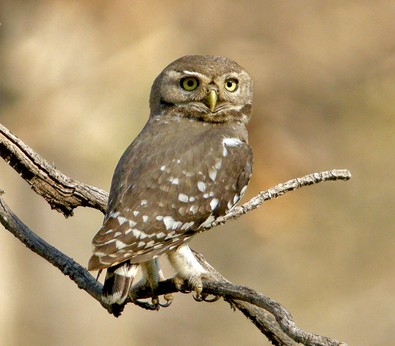}}
\hfil
\subfloat{\includegraphics[width=0.66in,height=0.55in]{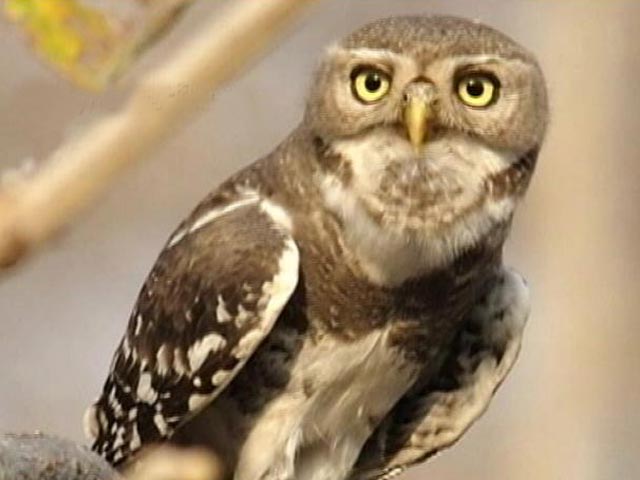}}
\hfil
\subfloat{\includegraphics[width=0.66in,height=0.55in]{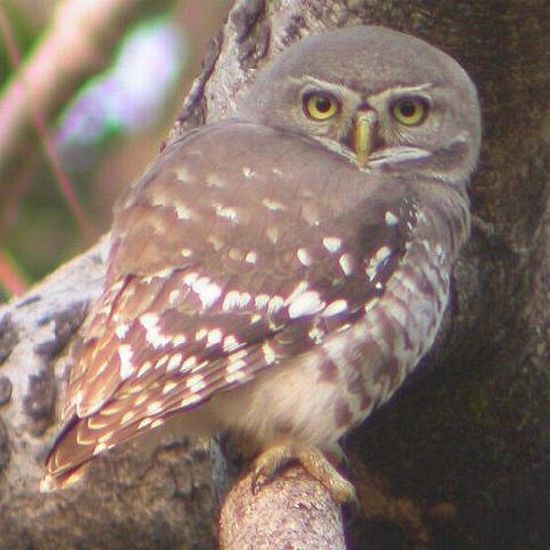}}
\hfil
\subfloat{\includegraphics[width=0.66in,height=0.55in]{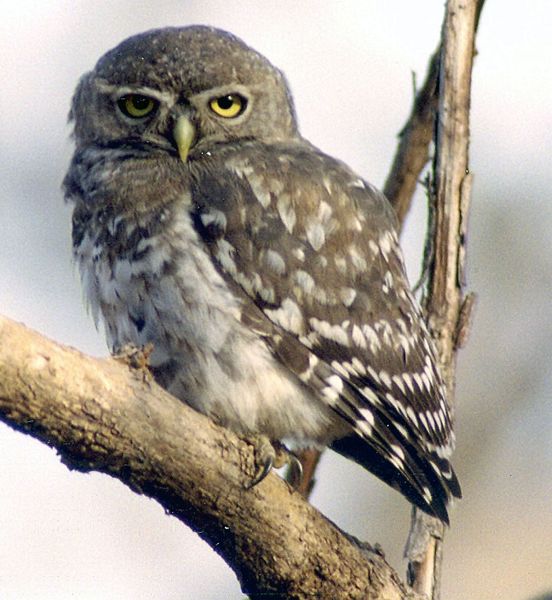}}
\hfil
\subfloat{\includegraphics[width=0.66in,height=0.55in]{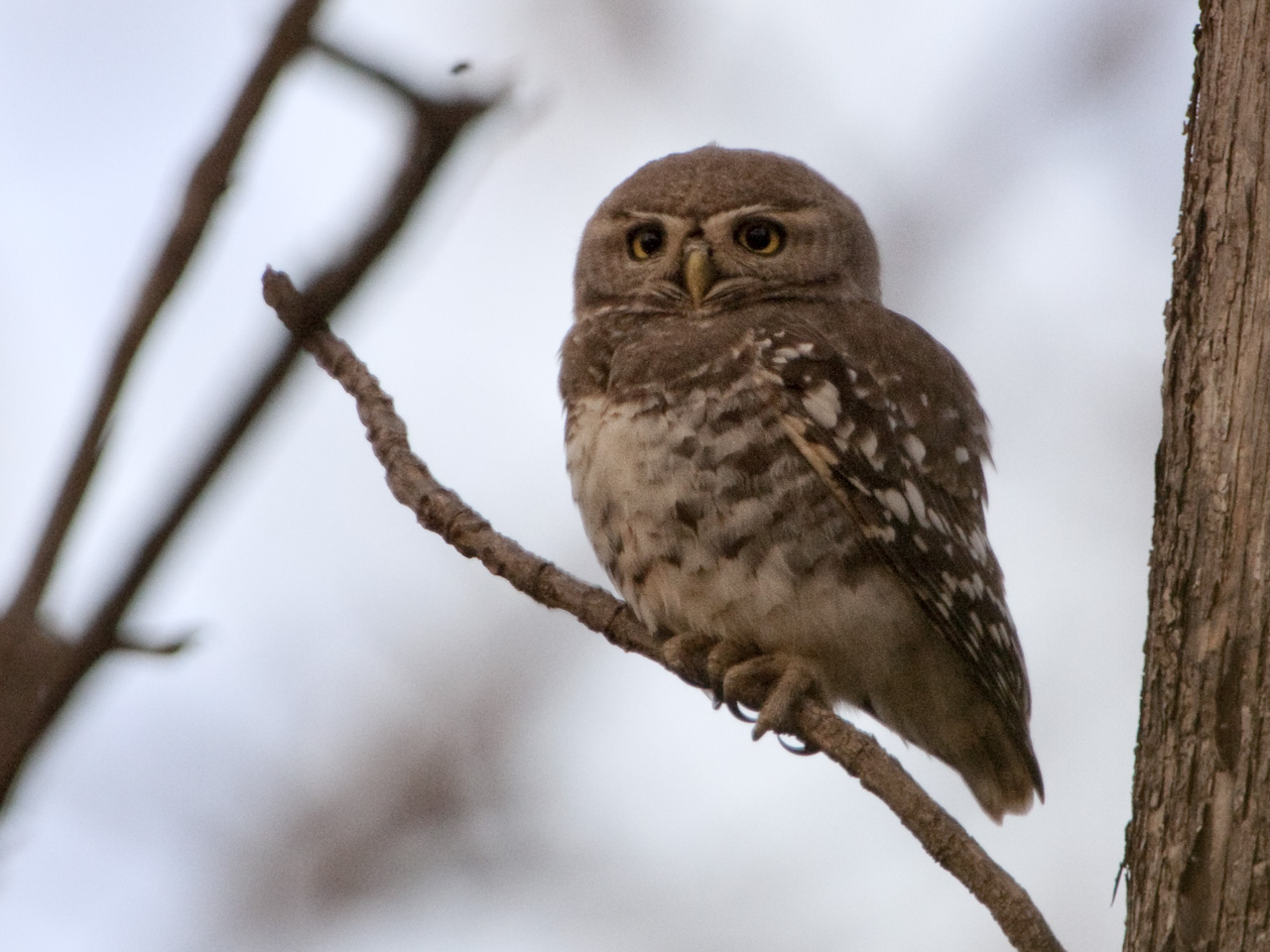}}
\\
Forest Owlet
\\
\subfloat{\includegraphics[width=0.66in,height=0.55in]{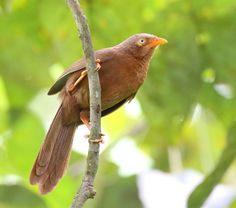}}
\hfil
\subfloat{\includegraphics[width=0.66in,height=0.55in]{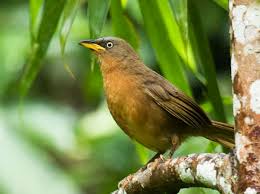}}
\hfil
\subfloat{\includegraphics[width=0.66in,height=0.55in]{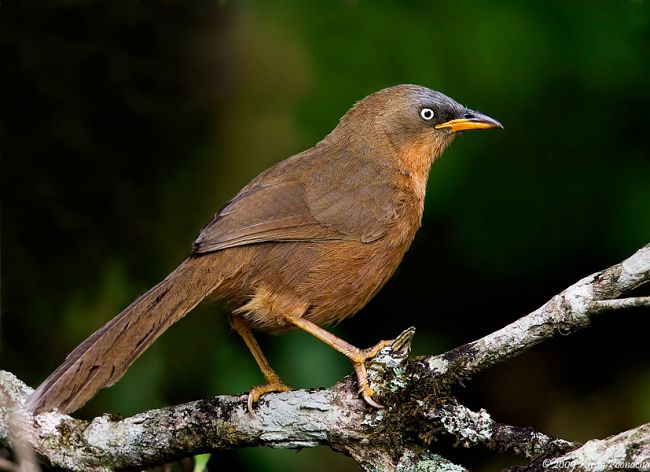}}
\hfil
\subfloat{\includegraphics[width=0.66in,height=0.55in]{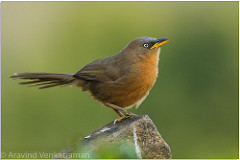}}
\hfil
\subfloat{\includegraphics[width=0.66in,height=0.55in]{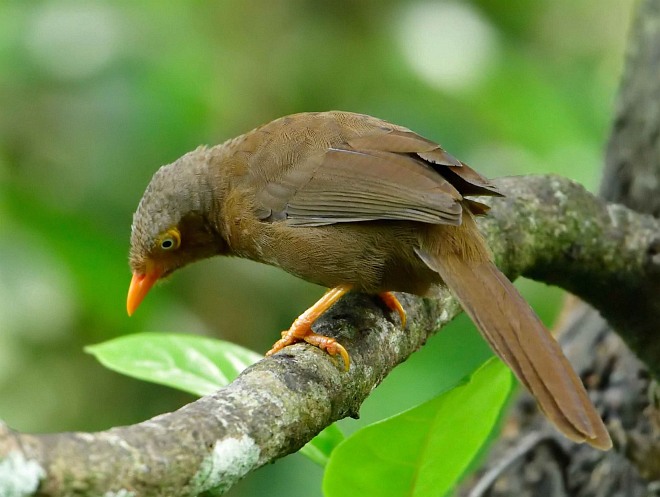}}
\\
Rufous Babbler
\\
\subfloat{\includegraphics[width=0.66in,height=0.55in]{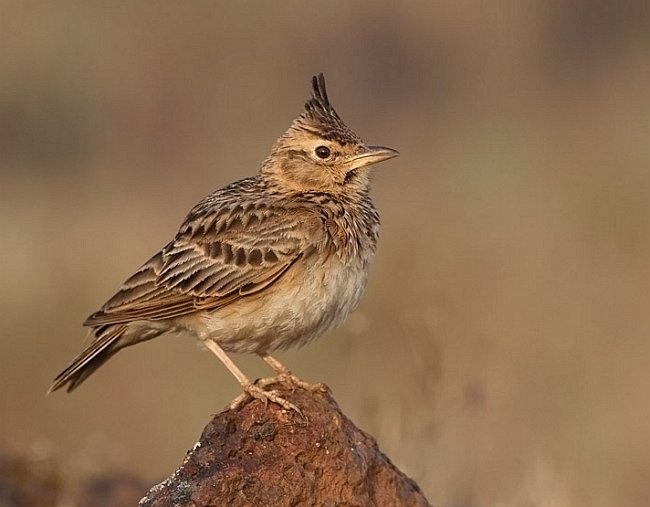}}
\hfil
\subfloat{\includegraphics[width=0.66in,height=0.55in]{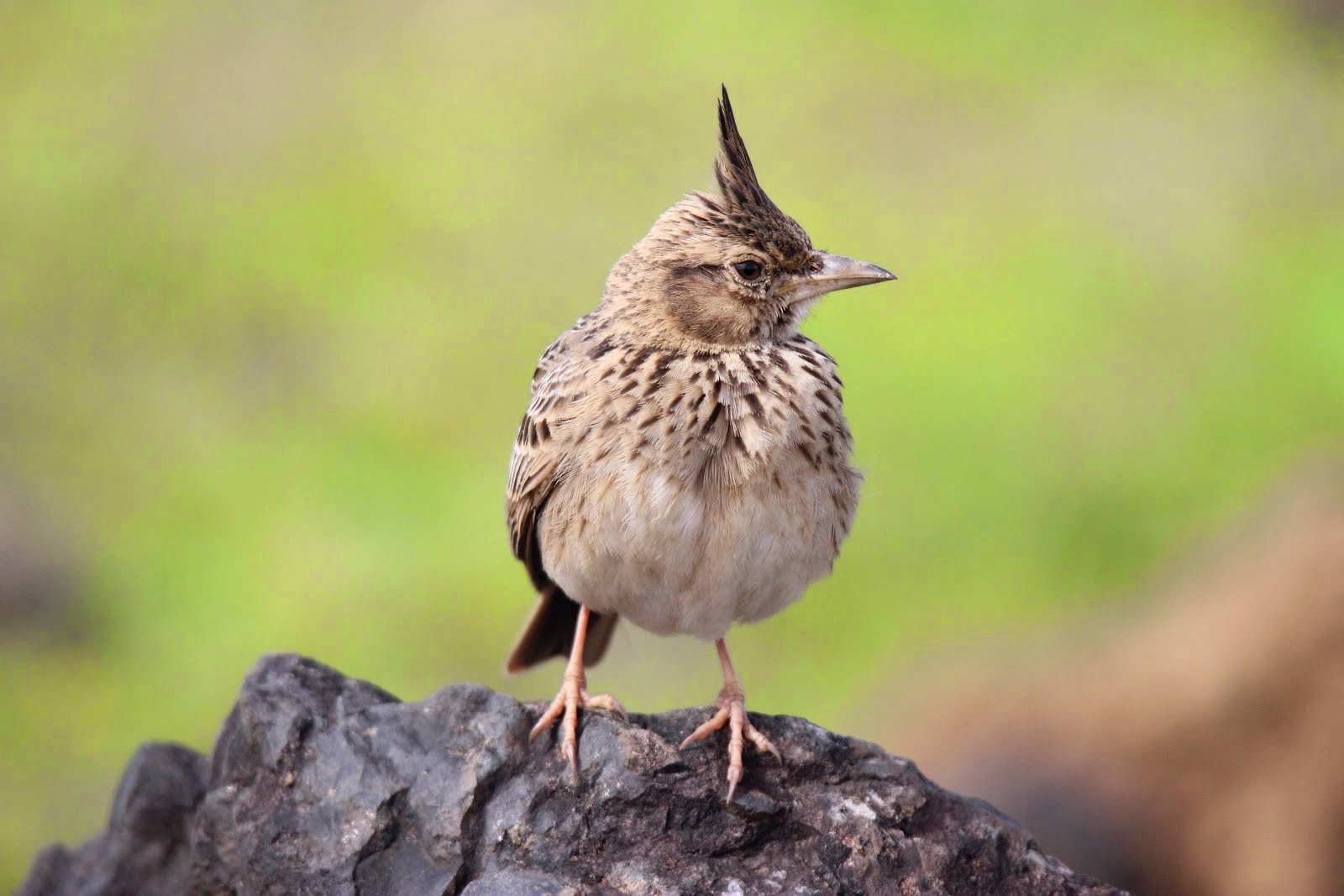}}
\hfil
\subfloat{\includegraphics[width=0.66in,height=0.55in]{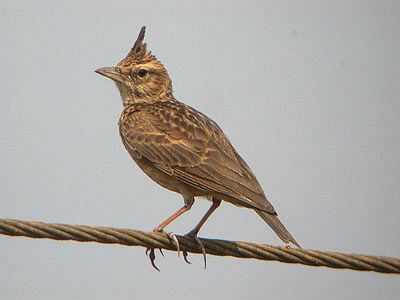}}
\hfil
\subfloat{\includegraphics[width=0.66in,height=0.55in]{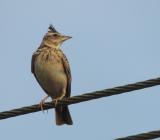}}
\hfil
\subfloat{\includegraphics[width=0.66in,height=0.55in]{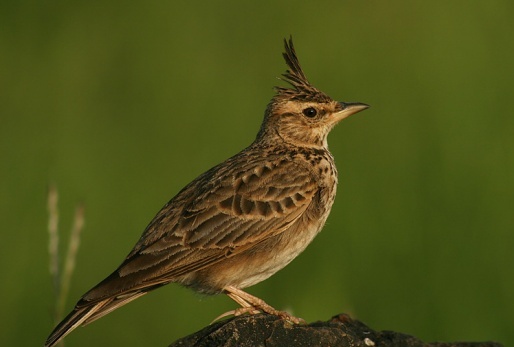}}
\\
Malabar Lark
\\
\subfloat{\includegraphics[width=0.66in,height=0.55in]{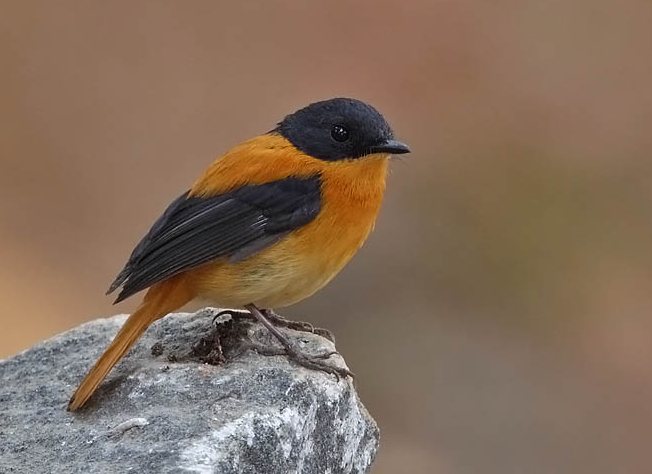}}
\hfil
\subfloat{\includegraphics[width=0.66in,height=0.55in]{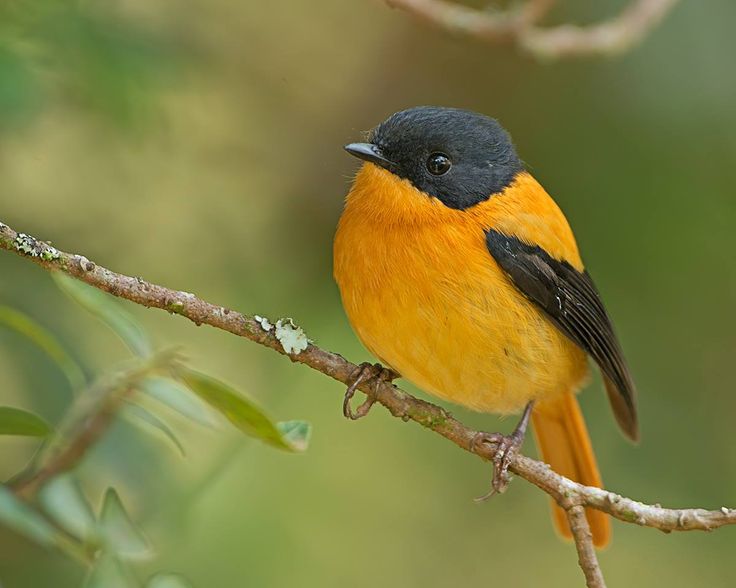}}
\hfil
\subfloat{\includegraphics[width=0.66in,height=0.55in]{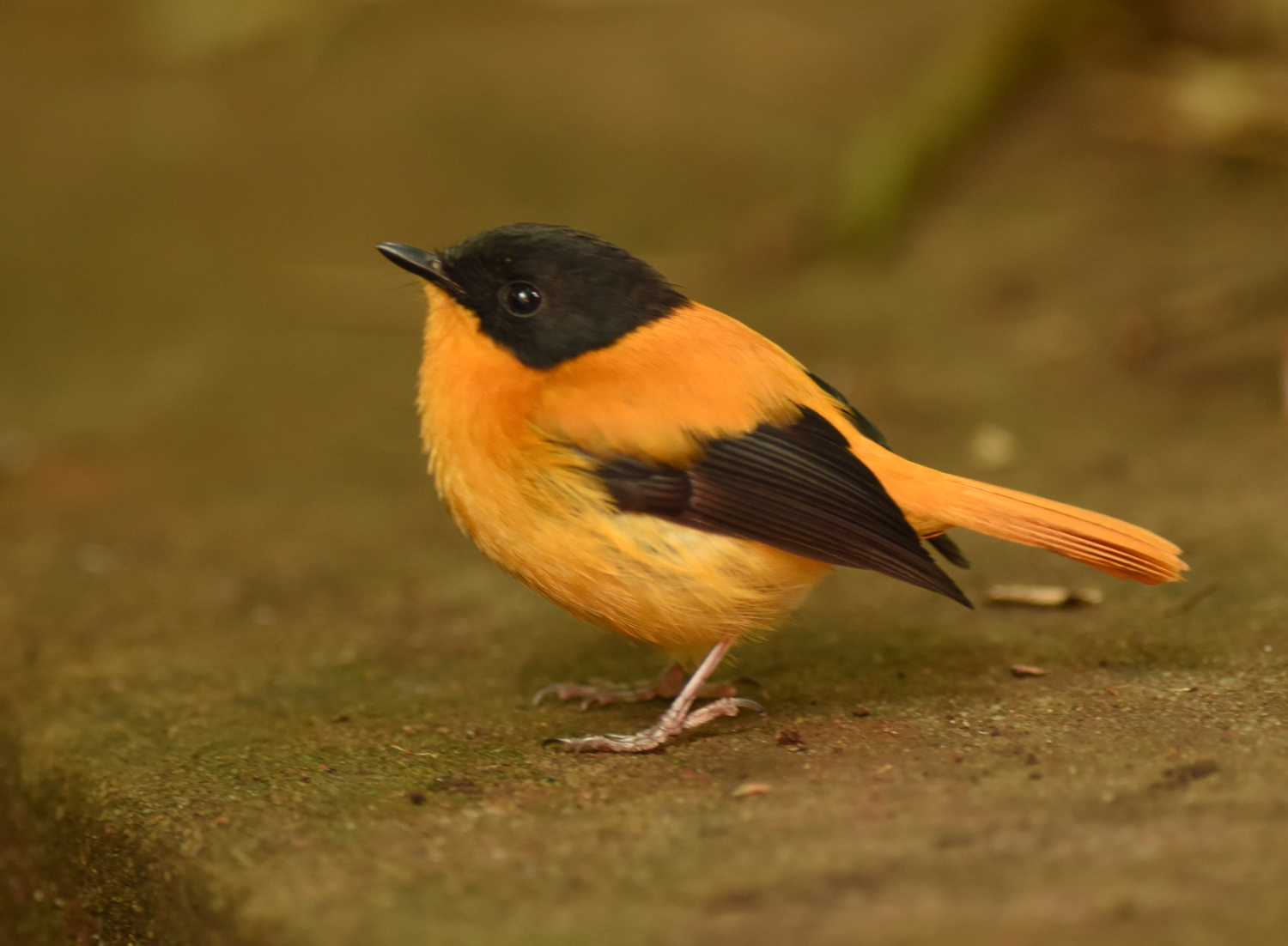}}
\hfil
\subfloat{\includegraphics[width=0.66in,height=0.55in]{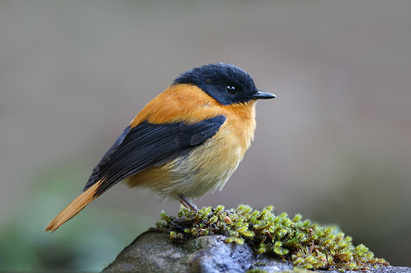}}
\hfil
\subfloat{\includegraphics[width=0.66in,height=0.55in]{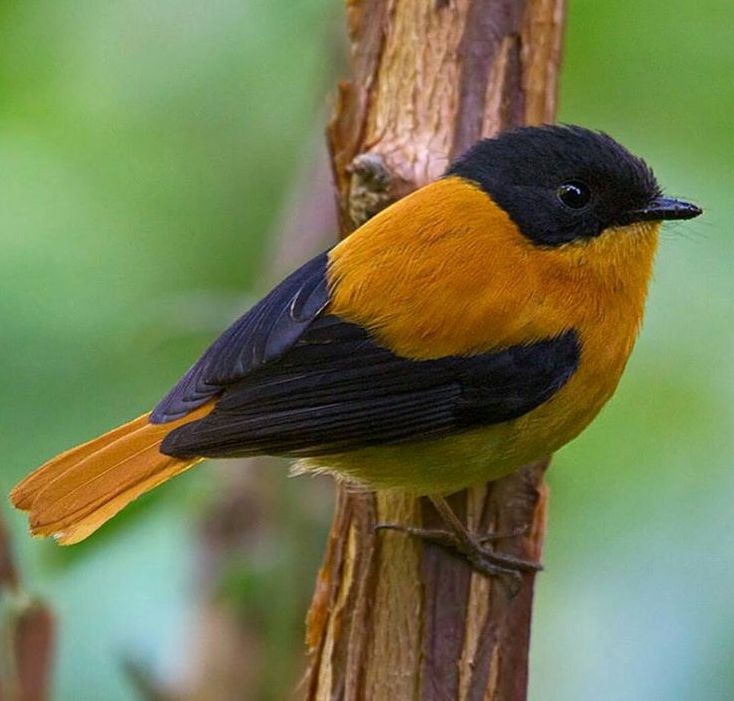}}
\\
Black and Orange Flycatcher

\caption{New IndBirds dataset: 8 classes; 100 images each}
\end{figure}

\subsection{Competing Classifiers}

We evaluate the performance of CoCoNet against popular recent methods both among collaborative representation classifiers (CRC) and deep convolutional neural networks (CNN), besides testing against constituent components in an ablation study. Among current CRC methods, we compare against the recent Probabilistic CRC (ProCRC) [10]. Among recent deep CNN models, we choose the popular Bilinear CNN [20],[21] and the very recent NTS-Net as the benchmark competitors [24]. 

Of course, there are a few more recent variants of ProCRC, like enhanced ProcCRC (EProCRC) [22], as well as BCNN, like improved BCNN [23]. However, we have deliberately used the vanilla versions because the aim is to establish a baselne evaluation in this work. For the same reason we also compare with the original CRC formulation [9] with VGG19 by Simonyan et al. [16] as the CNN backbone. \\

\textbf{Probabilistic CRC (ProCRC).} Cai \emph{et al}. [10] presented a probabilistic formulation (ProCRC) of the CRC method. Each of these probabilities are modeled by Gaussian exponentials and the probability of test image $y$ belonging to a class $k$ is expanded by chain rule using conditional probability. The final cost function for ProCRC is formulated as maximisation of the joint probability of the test image belonging to each of the possible classes as independent events. The final classification is performed by checking which class has the maximum likelihood.

\begin{equation}
J(\alpha,\lambda,\gamma)=\|y-X\alpha\|_2^2+\lambda\|\alpha\|_2^2+ \frac{\gamma}{K}\sum_{k=1}^{K}\|X\alpha-X_k\alpha_k\|_2^2
\end{equation}  \\




\textbf{Bilinear CNN.} Maji \emph{et al.} introduced the BCNN architecture for fine-grained visual recognition [20],[21]. These networks represent an image as a pooled outer product of features learned from two CNNs and encode localized feature interactions that are translationally invariant. B-CNN is a type of orderless texture representations that can be trained in an end-to-end manner.\\

\textbf{SOTA methods.} Besides the above, we also compare the performance of our method against several state-of-the-art approaches reported on the benchmark CUB dataset in ICCV 2019. These are Cross-X Learning [25], Deconstruction Construction Learning (DCL) [26], Trilinear Attention Sampling Network (TASN) [27], Selective Spars Sampling (S3N) [28] and Mixture of Granularity-Specific Experts (MGE-CNN) [29]. We see that our method compares favourably with these recent methods reported in results section.

\begin{table}[ht!]

\renewcommand{\arraystretch}{2}

\caption{{CUB 200-2011 Test Accuracy (\%)}}
\centering
\begin{tabular}{|p{2cm}|P{1.5cm}|P{1.5cm}|P{2.2cm}|}

\hline
& ImageNet \newline $\rightarrow$ CUB \newline \textbf{(1 stage)} & NABirds \newline $\rightarrow$ CUB \newline \textbf{(1 stage)} & ImageNet $\rightarrow$ NABirds$ \rightarrow$ CUB  \newline \textbf{(2 stage)} \\

\hline
Vgg19 & 71.9 $\pm$ 5.5 & 74.1 $\pm$ 5.7 & 77.5 $\pm$ 5.9 \\
\hline
Vgg19+CRC & 76.2 $\pm$ 5.6 & 79.0 $\pm$ 5.5 & 80.2 $\pm$ 5.9 \\
\hline
Vgg19+ProCRC & 79.3 $\pm$ 5.4 & 82.5 $\pm$ 5.5 & 83.8 $\pm$ 5.8 \\
\hline
\textbf{CoCoNet3} & \textbf{83.6 $\pm$ 5.5} & \textbf{87.4 $\pm$ 5.6} & \textbf{89.1 $\pm$ 5.6}  \\
\hline
\hline
\hline
Bilinear-CNN & 84.0 $\pm$ 5.3 & 85.7 $\pm$ 5.8 & 87.2 $\pm$ 5.5 \\
\hline
\bottomrule

\end{tabular}

\end{table}\textbf{}

\begin{table}[ht!]

\renewcommand{\arraystretch}{2}

\caption{{IndBirds Test Accuracy (\%)}}
\centering
\begin{tabular}{|p{2cm}|P{1.5cm}|P{1.5cm}|P{2.2cm}|}

\hline
& ImageNet \newline $\rightarrow$ IndBirds \newline \textbf{(1 stage)} & NABirds \newline $\rightarrow$ IndBirds \newline \textbf{(1 stage)} & ImageNet $\rightarrow$ NABirds$. \rightarrow$ IndBirds \textbf{(2 stage)} \\

\hline
Vgg19 & 76.2 $\pm$ 4.2 & 82.5 $\pm$ 4.7 & 84.8 $\pm$ 4.2 \\
\hline
Vgg19+CRC & 80.6 $\pm$ 4.4 & 86.3 $\pm$ 4.0 & 87.4 $\pm$ 4.4 \\
\hline
Vgg19+ProCRC & 84.0 $\pm$ 4.9 & 89.1 $\pm$ 4.1 & 91.0 $\pm$ 4.2 \\
\hline
\textbf{CoCoNet3} & \textbf{87.4 $\pm$ 4.3} & \textbf{92.9 $\pm$ 4.4} & \textbf{94.7 $\pm$ 4.5}  \\
\hline
\hline
\hline
Bilinear-CNN & 85.1 $\pm$ 4.7 & 88.6 $\pm$ 4.2 & 91.5 $\pm$ 4.3 \\
\hline
\bottomrule

\end{tabular}

\end{table}

\begin{table}[ht!]

\renewcommand{\arraystretch}{2}

\caption{{NZBirds Test Accuracy (\%)}}
\centering
\begin{tabular}{|p{2cm}|P{1.5cm}|P{1.5cm}|P{2.2cm}|}

\hline
& ImageNet $\rightarrow$ NZBirds \newline \textbf{(1 stage)} & NABirds $\rightarrow$ NZBirds \newline \textbf{(1 stage)} & ImageNet $\rightarrow$ NABirds $\rightarrow$ NZBirds \textbf{(2 stage)} \\

\hline
Vgg19 & 61.5 $\pm$ 5.0 & 63.7 $\pm$ 5.1 & 65.6 $\pm$ 5.7 \\
\hline
Vgg19+CRC & 63.9 $\pm$ 5.3 & 66.1 $\pm$ 5.5 & 68.7 $\pm$ 5.6 \\
\hline
Vgg19+ProCRC & 66.2 $\pm$ 5.5 & 71.3 $\pm$ 5.1 & 72.9 $\pm$ 5.8 \\
\hline
\textbf{CoCoNet3} & \textbf{71.8 $\pm$ 5.2} & \textbf{74.4 $\pm$ 5.2} & \textbf{77.2 $\pm$ 5.6} \\
\hline
\hline
\hline
Bilinear-CNN & 69.4 $\pm$ 5.6 & 71.8 $\pm$ 5.5 & 73.3 $\pm$ 5.0 \\
\hline
\bottomrule

\end{tabular}

\end{table}

\subsection{Experiments}

We train each of the three target datasets (CUB, NZBirds, IndBirds) through a combination of one stage and two stage transfer learning. For one stage transfer learning, two separate configurations have been used: 1) the network is pre-trained for general object recognition on ImageNet and then fine-tuned on the target dataset; 2) the network is pre-trained for bird recognition on the large north american bird dataset (NABirds) and then fine-tuned on the target dataset. For two stage training, the network is trained successively on ImageNet, NABirds and then the target dataset. Note that for pre-training, always the original architecture (VggNet) is used, CoCoNet only comes into play during fine-tuning. During both pre-training and fine-tuning, we start with 0.001 learning rate, but shift to 0.0001 once there is no change in loss anymore, keeping the total number of iterations/epochs constant at 1000 and using the Adam [17] optimiser. We investigate how the end-to-end formulation of CoCoNet fares in controlled experiments with competing configurations. We perform the same experiments using the original architecture (VggNet), then we observe change in accuracy with cascaded CNN+CRC and the end-to-end CoCoNet. We also further tabulate the results with cascaded CNN+ProCRC as well as Bilinear CNN [23]. For each dataset, images are resized to 128$\times$128 and experiments are conducted with five-fold cross validation. 

\subsection{Results and Analysis}

Percentage classification accuracies along with standard deviation are presented in Table 1 (CUB), Table 2 (IndBirds) and Table 3 (NZBirds). It may be readily observed from the tabulated results, that the proposed method overall outperforms its competitors, including the probabilistic CRC (ProCRC), bilinear CNN (BCNN). This performance is reflected across the three datasets and two stage transfer learning performs better than one stage for all classifiers. We compare with i) just a direct VggNet with the standard Softmax classfier and ii) VggNet as pre-trained feature extractor cascaded with a collaborative classifier (CRC) but not integrated. CoCoNet outperforms both of them as well as VggNet cascaded with probabilistic CRC (ProCRC), which is a more recent collaborative classifier. A qualitative example is presented in Fig. 3. \\



\textbf{Statistical Analysis.} We perform the Signed Binomial Test [30] to investigate the statistical significance of the improvement in performance of CoCoNet vs. BCNN, since these have the closest performances. The null hypothesis is that the two competitors are equally good, that is there is 50\% chance of each beating the other on any particular trial. For each of the three datasets (CUB, NZBirds and Indbirds), there are three transfer learning configurations  (two single stage and a double stage) and five-fold cross-validated results. Thus over the three datasets, in total we have 45 experiments of CoCoNet vs. each of its competitors. Out of these CoCoNet outperformed BCNN 33 times (that is 73.33\% of the trials). The signed binomial test yields that given the assumption that both methods are equally good, then the probability of CoCoNet outperforming BCNN in 73.33\% of the trials is has one -tail p-value of 0.0012 and 2-tail p-value of 0.0025. Considering a level of significance of $\alpha$ = 0.05, we have to apply the Bonferroni adjustment. We have 3 transfer learning protocols and 3 datasets: hence 9 combinations of experimental condition. So we divide the 5\% level of significance by 9 to get adjusted $\alpha$ = 0.0055. Since the p-values obtained in both cases is less than 0.0055, it may be concluded that the improvement in accuracy of the proposed method is statistically significant. \\

\begin{figure}[]
\centering

\subfloat[]{\includegraphics[width=1in,height=0.8in]{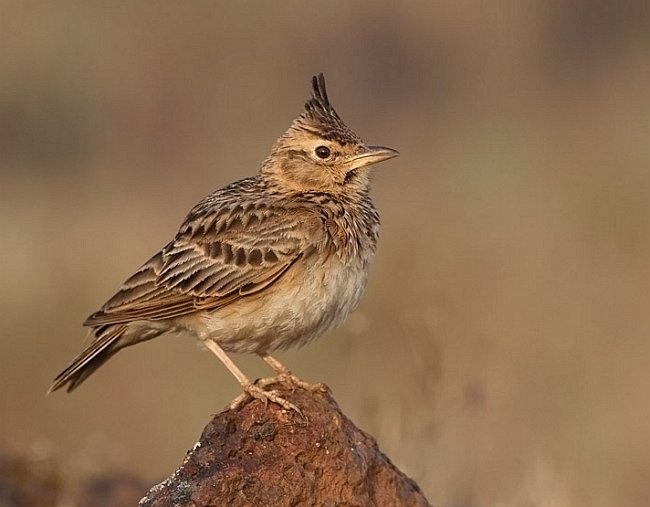}}
\hfil
\subfloat[]{\includegraphics[width=1in,height=0.8in]{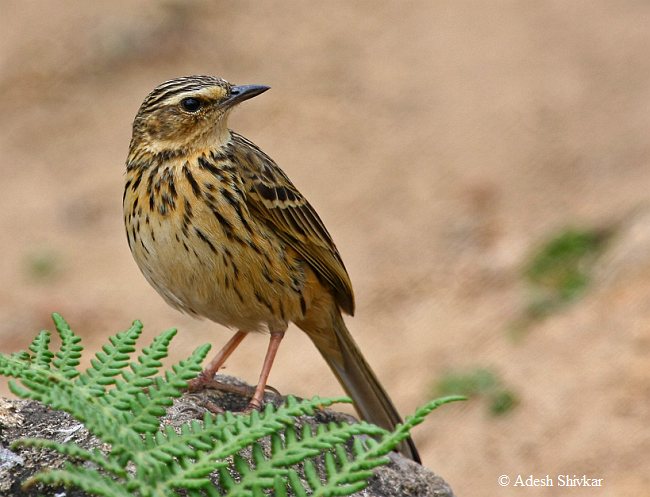}}
\hfil
\subfloat[]{\includegraphics[width=1in,height=0.8in]{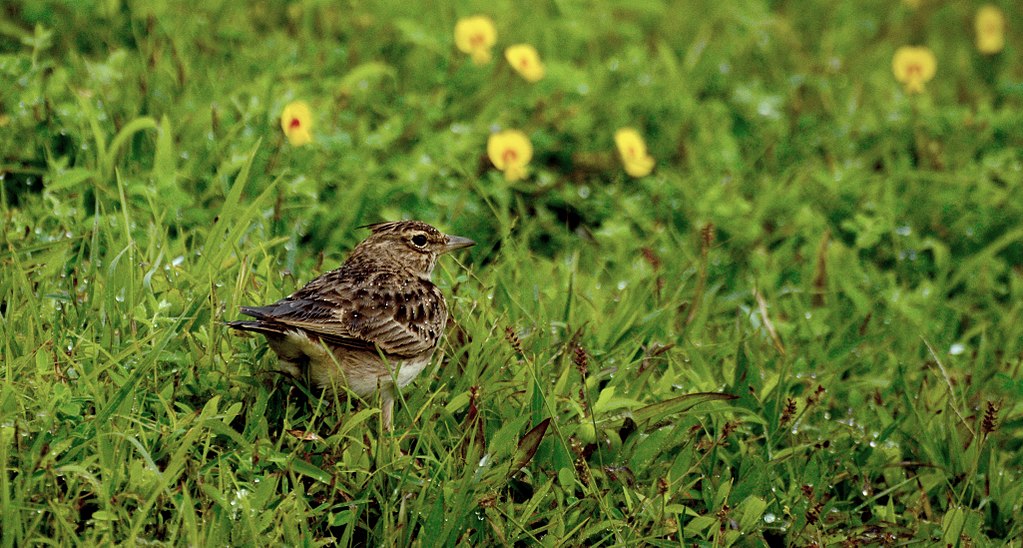}}
\hfil
\subfloat[]{\includegraphics[width=1in,height=0.8in]{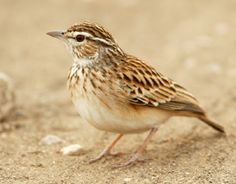}}
\hfil
\subfloat[]{\includegraphics[width=1in,height=0.8in]{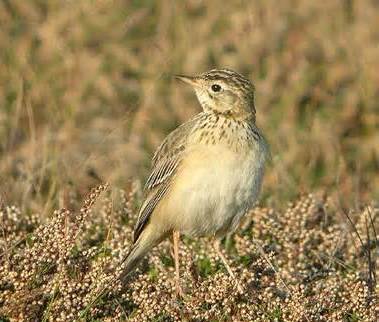}}
\hfil
\subfloat[]{\includegraphics[width=1in,height=0.8in]{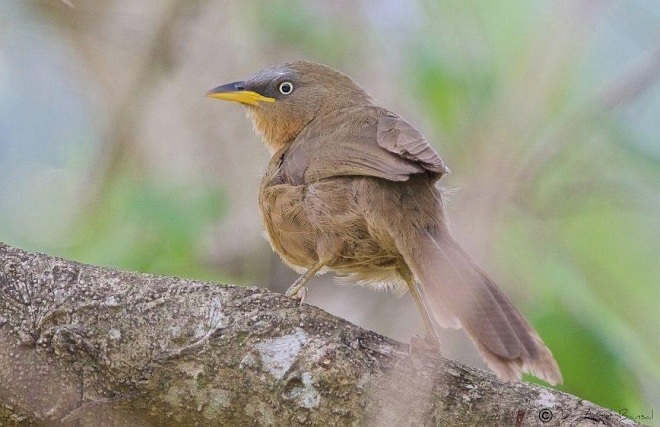}}

\caption{Classification and Misclassification Examples from the new IndBirds dataset: (a) Malabar Lark, (b) Nilgiri Pipit, (c) Malabar Lark, misclassified as Nilgiri Pipit by both proposed CoCoNet and competitors, due to obfuscation of the discriminating head crest. (d) Nilgiri Pipit with characteristic dark pattern on back (e) Front-facing image of Nilgiri Pipit with back patterns not visible. Correctly classified by proposed CoCoNet but misclassified by competitors as Rufous Babbler (f).}
\end{figure}

\begin{table*}
\begin{center}
\small
\begin{tabular}{|c|c|c|c|c|p{1.33cm}|p{1.33cm}|}
\hline
 \textbf{Cross-X [25]} &   \textbf{DCL [26]} & \textbf{TASN [27]}  &  \textbf{S3N [28]} & \textbf{MGE-CNN [29]} & \textbf{CoCoNet 1-stage} & \textbf{CoCoNet 2-stage} \\
\hline\hline
88.5 & 87.7 & 88.5 & 87.9 & 87.8 & \textbf{87.4} & \textbf{89.1} \\
 \hline
\end{tabular}
\end{center}
\caption{CoCoNet with 2 stage transfer learning against state-of-the-art on CUB dataset}
\end{table*}


\textbf{SOTA Comparison.} In Table 4, we compare the performance of our method directly against several state-of-the-art methods from ICCV 2019 that reported results on the CUB dataset. These are Cross-X Learning [25], Deconstruction Construction Learning (DCL) [26], Trilinear Attention Sampling Network (TASN) [27], Selective Spars Sampling (S3N) [28] and Mixture of Granularity-Specific Experts (MGE-CNN) [29]. It can be noted that our method outperforms them for 2-stage transfer learning and for 1-stage transfer learning, the results are comparable.

\section{Conclusion}

We present an end-to-end collaborative convolutional network (CoCoNet) architecture for fine-grained visual recognition of bird species with limited samples. The new architecture adds a collaborative representation which adds an additional level of optimization based on collaboration of images across classes, the information is then back-propagated to update CNN weights in an end-to-end fashion. This collaborative representation exploits the fine-grained nature of the data better with fewer training images. The proposed network is evaluated for the task of fine-grained bird species recognition, but the method is general enough to perform other fine-grained classification tasks like detection of rare pathology from medical images. The other major advantage is that most existing CNN architectures can be easily restructured into the proposed configuration. We also present a new fine-grained benchmark dataset with images of endemic Indian birds, and report results on it. Results show that the proposed algorithm performs better than its constituent parts and several state-of-the-art competitors. \\

\bibliographystyle{ieee}

\end{document}